\documentclass[sigconf]{acmart}

\usepackage{booktabs} % For formal tables

\setcopyright{rightsretained}
\copyrightyear{2020}
\acmYear{2020}
\setcopyright{rightsretained}
\acmConference[GECCO '20 Companion]{Genetic and Evolutionary Computation Conference Companion}{July 8--12, 2020}{Cancún, Mexico}
\acmBooktitle{Genetic and Evolutionary Computation Conference Companion (GECCO '20 Companion), July 8--12, 2020, Cancún, Mexico}\acmDOI{10.1145/3377929.3389901}
\acmISBN{978-1-4503-7127-8/20/07}

\acmPrice{15.00}

\begin{document}
\title{Adaptive Reinforcement Learning through Evolving Self-Modifying Neural Networks}
%\titlenote{Produces the permission %block, and
%  copyright information}
%\subtitle{Subtitle}
%\subtitlenote{The full version of %the author's guide is available as
%  \texttt{acmart.pdf} document}

%\author{Anonymous authors}
%\affiliation{
%  \institution{Paper under double-blind review}
%}

\author{Samuel Schmidgall}
\affiliation{
  \institution{George Mason University}
}
\email{sschmidg@gmu.edu}

\begin{abstract}
The adaptive learning capabilities seen in biological neural networks are largely a product of the self-modifying behavior emerging from online plastic changes in synaptic connectivity. Current methods in Reinforcement Learning (RL) only adjust to new interactions after reflection over a specified time interval, preventing the emergence of online adaptivity. Recent work addressing this by endowing artificial neural networks with neuromodulated plasticity have been shown to improve performance on simple RL tasks trained using backpropagation, but have yet to scale up to larger problems. Here we study the problem of meta-learning in a challenging quadruped domain, where each leg of the quadruped has a chance of becoming unusable, requiring the agent to adapt by continuing locomotion with the remaining limbs. Results demonstrate that agents evolved using self-modifying plastic networks are more capable of adapting to complex meta-learning learning tasks, even outperforming the same network updated using gradient-based algorithms while taking less time to train.

%"while avoiding the added drawbacks of training recurrent connections"

%\footnote{This is an abstract footnote}
\end{abstract}


\maketitle

\section{Introduction \& Related Work}
% if task not able to be separated or even if not introduced to it yet

The brain's active self-modifying behavior plays an important role in its effectiveness for continual adaptation and learning in dynamic environments. Furthermore, evolution has led to the design of both the underlying neural connectivity as well as the framework for directing neuromodulated plasticity, the structure from which short-term synaptic self-modification occurs. However, the most common methods from which current AI are trained contradicts this way of learning. Consequently, modern training methods render AI incapable of online adaptation, only performing well on the tasks that they were trained on. Even slight deviations from the original simulated environment might be catastrophic for the agent's performance.

To address this problem, recent literature in meta-learning aim to optimize toward an initial set of parameters that enable rapid learning over a specified set of tasks, such as Model-Agnostic Meta-Learning (MAML) ~\cite{DBLP:journals/corr/FinnAL17}. Another set of methods utilize fast and slow-weights in neural networks through a non-trainable Hebbian learning-based associative memory ~\cite{DBLP:journals/corr/abs-1803-10049}. Building off of this, differential neuromodulation ~\cite{Miconi2019BackpropamineTS} proposes a way to augment traditional artificial neural networks with fast- and slow-weights, where the fast-weights are modified through the addition of neuromodulated plasticity that is trainable using backpropagated gradients.
 
The work presented in this paper both demonstrates that self-modifying neural networks are capable of solving complex learning tasks in dynamic environments and poses Evolutionary Strategies as the natural choice for developing such networks. Previous work using neurmodulated plasticity ~\cite{Miconi2019BackpropamineTS}[5] only experimented on simple problems, and only considered optimization through backpropagating gradients. Here we show evidence toward the applicability of evolved neuromodulated plasticity in the high-dimensional continuous control problem, Crippled-Ant, requiring both precise motor skills and adaptivity. %Evolutionary Strategies is proposed as the preferred optimization framework through performance and time comparisons on the CrippledAnt benchmark.

%OpenAI-ES [3] is used to optimize 

\section{Methods}
The approach presented in this work compares a traditional neural network architecture against one with self-modifying synaptic connectivity, where the changes in connectivity are modulated by a learned set of parameters. Performance comparisons are made between policy gradient algorithm Proximal Policy Optimization ~\cite{DBLP:journals/corr/SchulmanWDRK17} and a simplified version of Natural Evolutionary Strategies ~\cite{salimans2017evolution}, which, for simplicity, will be referred to as OpenAI-ES for the duration of this paper. 
\subsection{Differential Neuromodulation}
Within the differential neuromodulation framework, the weights along with the plasticity of each connection are optimized: % ONLY COMPARE TIME USING NEUROMODULATION
\begin{equation}
x_{t} = \phi ((w + \alpha H_{t})x_{t-1})
\end{equation}
\begin{equation}
H_{t+1} = H_{t} + M(x_{t})x_{t-1}x_{t}
\vspace{1.3mm}
\end{equation}
where $x_{t}$ is the output of a layer of neurons at time $t$, $\phi$ is a nonlinear activation function, $w$ is the set of traditional non-plastic weights, and $\alpha$ is the plasticity coefficient that scales the magnitude of the plastic component of each connection. The plastic component at timestep $t$ is represented by $H_{t}$, which accumulates the modulated product of pre- and post-synaptic activity between the respective layers. Here, plasticity is modulated through a network learned neuromodulatory signal $M(x_{t})$, which be represented by a variety of functions, but in this work is represented by a single-layer feedforward neural network. $H_{t}$ is generally clipped between -$\omega$ and $\omega$, with $\omega$ = 1 in this experiment.

\begin{figure}
\includegraphics[scale=0.207]{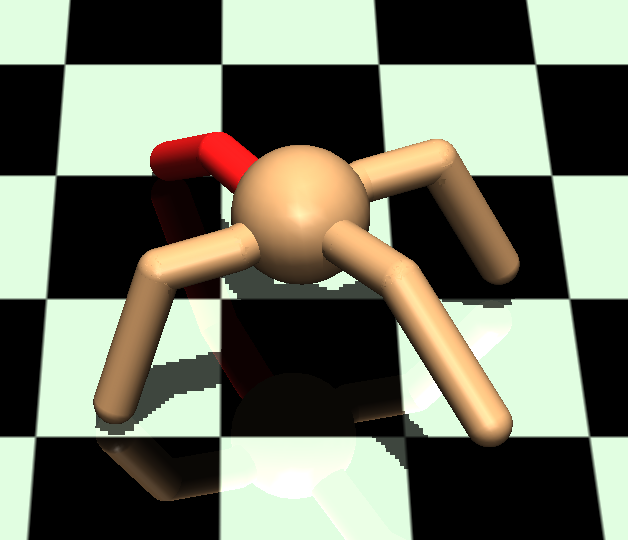}
\includegraphics[scale=0.2]{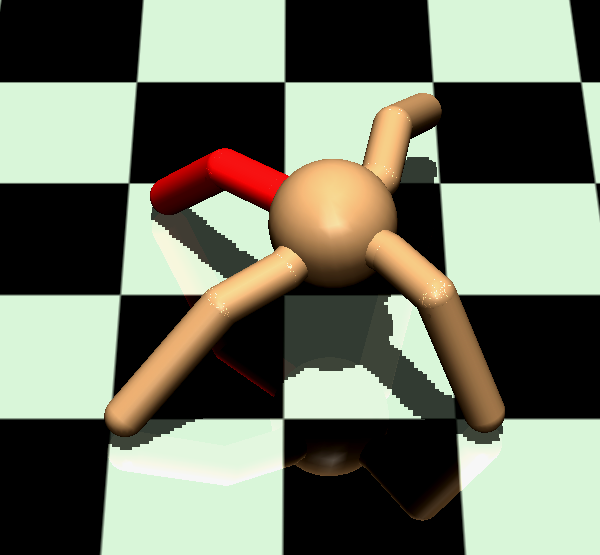}
\caption{Adaptive locomotion. \textmd{In the Crippled-Ant Environment, a limb is chosen at random to be disabled (in red) requiring the agent to adapt its gait using the remaining limbs.}}

\end{figure}

\subsection{OpenAI-ES}
Starting with an initial zero-vector $\theta_t$, the OpenAI-ES algorithm generates N population samples of random noise vectors $v_{t,i}\sim \mathcal{N}(0, \sigma)$ and uses them to create population individuals $\theta_t + v_{t,i}$. The fitness of each individual is evaluated over the course of a lifetime through an environment defined reward, $r_{t,i}$. Such reward is often center-ranked to prevent early local optima \cite{salimans2017evolution}. Using the corresponding rewards, parameters are updated with Stochastic Gradient Descent (SGD) as follows:

\begin{equation}
\theta_{t+1} = \theta_{t}+
\alpha\dfrac{1}{N\sigma^{2}}\sum_{n=1}^{N}v_{t,i}r_{t,i}
\end{equation}
\vspace{1.3mm}

OpenAI-ES was chosen because it has been shown to be competitive with and exhibit better exploration behavior than both DQN and A3C on difficult RL benchmarks ~\cite{salimans2017evolution}. While OpenAI-ES is less sample-efficient than these other methods, it is better structured for distributed computing and allows a shorter wall-clock training time. Additionally, due to not requiring back-propagation of error gradients, the required wall-clock training time is further significantly reduced for optimization over networks involving recurrence, such as the neuromodulated plasticity used in our experiments.

\subsection{Crippled-Ant Environment}
The meta-learning capabilities of the neural network in this paper are evaluated on a high-dimensional continuous control environment, Crippled-Ant ~\cite{DBLP:journals/corr/abs-1803-11347}. The environment begins with a 12-jointed quadruped aiming to attain the highest possible velocity in a limited amount of time (Figure 1). The environment takes direct joint torque for each of the 12 joints as input. The state is represented as a 111 dimensional vector containing relative angles and velocities for each joint, as well as information about external forces acting on the quadruped. At the beginning of each session, a leg is randomly selected to be crippled on the quadrupedal robot, rendering it fully unusable. This environment was chosen because this modification causes significant change in the action dynamics, requiring gait adaptation throughout the course of each run.

%is rewarded based on its current velocity

\section{Results \& Discussion}
Evaluation of performance is averaged over 100 episodes from 5 fully trained models for each algorithm during the testing phase to ensure accurate measurement. Each algorithm is trained using the default hyper-parameters from their respective papers. OpenAI-ES was compared against a policy gradient algorithm often used in continuous control problems, Proximal Policy Optimization (PPO). Both of these algorithms were also compared using fixed weights and differential self-modifying ones. The experimental results demonstrate that self-modifying networks trained through Evolutionary Strategies consistently outperform networks without such augmentation trained using OpenAI-ES and PPO, as well as self-modifying networks using PPO. Total training time for the self-modifying OpenAI-ES averaged around \textbf{214.8} minutes, and \textbf{968.8} minutes for the self-modifying PPO running on a standard 6-core CPU. Future work involves experimenting with new types of neuromodulation, as well as understanding the full capabilities of such networks.

\begin{figure}
\includegraphics[width=\linewidth]{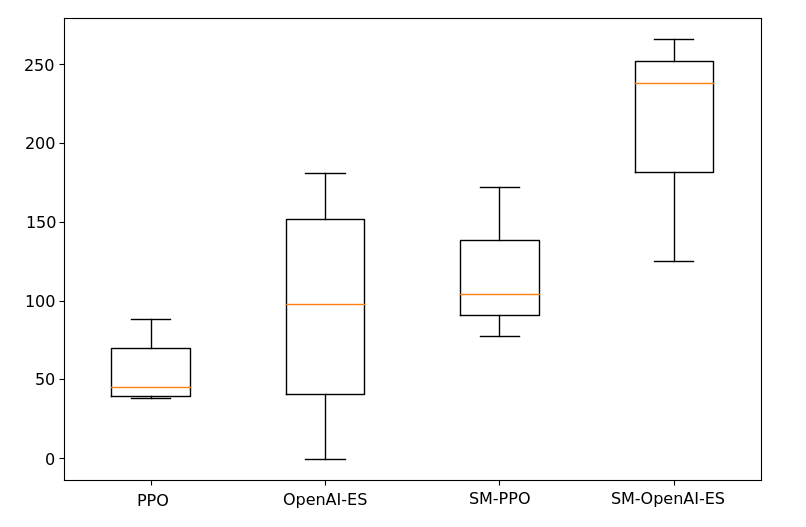}
\caption{Performance Comparison on Crippled-Ant Environment \textmd{Performance of each policy is measured for self-modifying (SM-) and traditional neural networks trained using Proximal Policy Optimization and OpenAI-ES.}}

\end{figure}

%\appendix
%Appendix A

%~\cite{Miconi2019BackpropamineTS}
%~\cite{salimans2017evolution}

%[1] Jack W Rae, Chris Dyer, Peter Dayan, and Timothy P Lillicrap. Fast parametric learning with activation memorization. arXiv preprint arXiv:1803.10049, 2018.

%[2] Chelsea Finn, Pieter Abbeel, and Sergey Levine. Model-agnostic meta-learning for fast adaptation of deep networks. International Conference on Machine Learning, 2017.

%[3] Ignasi Clavera, Anusha Nagabandi, Simin Liu, Ronald S. Fearing, Pieter Abbeel, Sergey Levine, and Chelsea Finn. Learning to adapt in dynamic, real-world environments through meta-reinforcement learning. In International Conference on Learning Representations, 2019. 

%[4] Thomas Miconi, Aditya Rawal, Jeff Clune, and Kenneth O Stanley. Backpropamine: training self-modifying neural networks with differentiable neuromodulated plasticity. International Conference on Learning Representations, 2019.

%[n] T. Salimans, J. Ho, X. Chen, and I. Sutskever. Evolution strategies as a scalable alternative to reinforcement learning. arXiv preprint arXiv:1703.03864, 2017.

% ~~ give reasoning for using ES
% Furthermore, the time required to train such networks is drastically increased due to the recurrent nature of the network structure. Evolutionary Strategies do not suffer from the same issue since performance is evaluated over entire series of interactions rather than each individual state action pair itself.

%\bibliographystyle{ACM-Reference-Format}
%\bibliography{sample-bibliography} 

%\bibliography{sample-bibliography} 

\bibliographystyle{ACM-Reference-Format}
\bibliography{sample-bibliography} 


\begin{thebibliography}{6}

%%% the Web address.

\ifx \showCODEN    \undefined \def \showCODEN     #1{\unskip}     \fi
\ifx \showDOI      \undefined \def \showDOI       #1{#1}\fi
\ifx \showISBNx    \undefined \def \showISBNx     #1{\unskip}     \fi
\ifx \showISBNxiii \undefined \def \showISBNxiii  #1{\unskip}     \fi
\ifx \showISSN     \undefined \def \showISSN      #1{\unskip}     \fi
\ifx \showLCCN     \undefined \def \showLCCN      #1{\unskip}     \fi
\ifx \shownote     \undefined \def \shownote      #1{#1}          \fi
\ifx \showarticletitle \undefined \def \showarticletitle #1{#1}   \fi
\ifx \showURL      \undefined \def \showURL       {\relax}        \fi
% invisible to TeX
\providecommand\bibfield[2]{#2}
\providecommand\bibinfo[2]{#2}
\providecommand\natexlab[1]{#1}
\providecommand\showeprint[2][]{arXiv:#2}

\bibitem[\protect\citeauthoryear{Clavera, Nagabandi, Fearing, Abbeel, Levine,
  and Finn}{Clavera et~al\mbox{.}}{2018}]%
        {DBLP:journals/corr/abs-1803-11347}
\bibfield{author}{\bibinfo{person}{Ignasi Clavera}, \bibinfo{person}{Anusha
  Nagabandi}, \bibinfo{person}{Ronald~S. Fearing}, \bibinfo{person}{Pieter
  Abbeel}, \bibinfo{person}{Sergey Levine}, {and} \bibinfo{person}{Chelsea
  Finn}.} \bibinfo{year}{2018}\natexlab{}.
\newblock \showarticletitle{Learning to Adapt: Meta-Learning for Model-Based
  Control}.
\newblock \bibinfo{journal}{{\em CoRR\/}}  \bibinfo{volume}{abs/1803.11347}
  (\bibinfo{year}{2018}).
\newblock
\showeprint[arxiv]{1803.11347}
\showURL{%
\url{http://arxiv.org/abs/1803.11347}}


\bibitem[\protect\citeauthoryear{Finn, Abbeel, and Levine}{Finn
  et~al\mbox{.}}{2017}]%
        {DBLP:journals/corr/FinnAL17}
\bibfield{author}{\bibinfo{person}{Chelsea Finn}, \bibinfo{person}{Pieter
  Abbeel}, {and} \bibinfo{person}{Sergey Levine}.}
  \bibinfo{year}{2017}\natexlab{}.
\newblock \showarticletitle{Model-Agnostic Meta-Learning for Fast Adaptation of
  Deep Networks}.
\newblock \bibinfo{journal}{{\em CoRR\/}}  \bibinfo{volume}{abs/1703.03400}
  (\bibinfo{year}{2017}).
\newblock
\showeprint[arxiv]{1703.03400}
\showURL{%
\url{http://arxiv.org/abs/1703.03400}}


\bibitem[\protect\citeauthoryear{Miconi, Rawal, Clune, and Stanley}{Miconi
  et~al\mbox{.}}{2019}]%
        {Miconi2019BackpropamineTS}
\bibfield{author}{\bibinfo{person}{Thomas Miconi}, \bibinfo{person}{Aditya
  Rawal}, \bibinfo{person}{Jeff Clune}, {and} \bibinfo{person}{Kenneth~O.
  Stanley}.} \bibinfo{year}{2019}\natexlab{}.
\newblock \showarticletitle{Backpropamine: training self-modifying neural
  networks with differentiable neuromodulated plasticity}. In
  \bibinfo{booktitle}{{\em ICLR}}.
\newblock


\bibitem[\protect\citeauthoryear{Rae, Dyer, Dayan, and Lillicrap}{Rae
  et~al\mbox{.}}{2018}]%
        {DBLP:journals/corr/abs-1803-10049}
\bibfield{author}{\bibinfo{person}{Jack~W. Rae}, \bibinfo{person}{Chris Dyer},
  \bibinfo{person}{Peter Dayan}, {and} \bibinfo{person}{Timothy~P. Lillicrap}.}
  \bibinfo{year}{2018}\natexlab{}.
\newblock \showarticletitle{Fast Parametric Learning with Activation
  Memorization}.
\newblock \bibinfo{journal}{{\em CoRR\/}}  \bibinfo{volume}{abs/1803.10049}
  (\bibinfo{year}{2018}).
\newblock
\showeprint[arxiv]{1803.10049}
\showURL{%
\url{http://arxiv.org/abs/1803.10049}}


\bibitem[\protect\citeauthoryear{Salimans, Ho, Chen, Sidor, and
  Sutskever}{Salimans et~al\mbox{.}}{2017}]%
        {salimans2017evolution}
\bibfield{author}{\bibinfo{person}{Tim Salimans}, \bibinfo{person}{Jonathan
  Ho}, \bibinfo{person}{Xi Chen}, \bibinfo{person}{Szymon Sidor}, {and}
  \bibinfo{person}{Ilya Sutskever}.} \bibinfo{year}{2017}\natexlab{}.
\newblock \bibinfo{title}{Evolution Strategies as a Scalable Alternative to
  Reinforcement Learning}.
\newblock   (\bibinfo{year}{2017}).
\newblock
\showeprint[arxiv]{stat.ML/1703.03864}


\bibitem[\protect\citeauthoryear{Schulman, Wolski, Dhariwal, Radford, and
  Klimov}{Schulman et~al\mbox{.}}{2017}]%
        {DBLP:journals/corr/SchulmanWDRK17}
\bibfield{author}{\bibinfo{person}{John Schulman}, \bibinfo{person}{Filip
  Wolski}, \bibinfo{person}{Prafulla Dhariwal}, \bibinfo{person}{Alec Radford},
  {and} \bibinfo{person}{Oleg Klimov}.} \bibinfo{year}{2017}\natexlab{}.
\newblock \showarticletitle{Proximal Policy Optimization Algorithms}.
\newblock \bibinfo{journal}{{\em CoRR\/}}  \bibinfo{volume}{abs/1707.06347}
  (\bibinfo{year}{2017}).
\newblock
\showeprint[arxiv]{1707.06347}
\showURL{%
\url{http://arxiv.org/abs/1707.06347}}


\end{thebibliography}

\end{document}